\documentclass{ecai}
\usepackage{times}
\usepackage{graphicx}
\usepackage{latexsym}
\usepackage{subfiles}
\usepackage{hyperref}
\usepackage{subcaption}
\usepackage{float}
\usepackage{amsmath}
\usepackage{algorithm}
\usepackage{algorithmic}
\usepackage{breqn}
\usepackage{adjustbox}
\usepackage{tikz}
\usepackage{dsfont}

\begin{document}

\title{Building a Multi-domain Neural Machine Translation Model using Knowledge Distillation}

\author{Idriss MGHABBAR\institute{BNP Paribas,
France, email: idriss.mg@gmail.com}  \and Pirashanth RATNAMOGAN \institute{BNP Paribas,
France, email: pirashanth.ratnamogan@bnpparibas.com} }

\maketitle
\bibliographystyle{ecai}

\begin{abstract}
Lack of specialized data makes building a multi-domain neural machine translation tool challenging. Although emerging literature dealing with low resource languages starts to show promising results, most state-of-the-art models used millions of sentences. Today, the majority of multi-domain adaptation techniques are based on complex and sophisticated architectures that are not adapted for real-world applications. So far, no scalable method is performing better than the simple yet effective mixed-finetuning, i.e finetuning a generic model with a mix of all specialized data and generic data. In this paper, we propose a new training pipeline where knowledge distillation and multiple specialized teachers allow us to efficiently finetune a model without adding new costs at inference time. Our experiments demonstrated that our training pipeline allows improving the performance of multi-domain translation over finetuning in configurations with 2, 3, and 4 domains by up to 2 points in BLEU.

\end{abstract}

\section{Introduction} 
Since statistical machine translation, domain adaptation has always been a field of machine translation that has interested business and academic research. Indeed, model specialization methods now allow building models that perform better on specialized domains such as finance, marketing or science. \\

In this paper, we try to challenge the state of the art to solve a critical scalability problem for real-world applications: building an efficient and effective multi-domain neural machine translation model that will ensure high-quality translation in a large set of domains. Indeed, most of the companies have multiple services, such as legal, marketing or accounting, that require multiple specialized translation models. However deploying several single domain models in production is not scalable as it introduces storage costs and complexity. \\

Today, most of the solutions to tackle the multi-domain specialization problem suffers from a lack of effectiveness as they rely either on ensembling, reranking or prior classification of input sentences when translating. One of the key advantages of our work is that the original architecture is not altered and that there is no added complexity or parameter increase. \\

To do so, we leverage the well-known knowledge distillation to tackle the multi-domain neural machine translation problem. Knowledge distillation has been used as a compression technique where light models are learning to mimic cumbersome models \cite{distillation}. Very recently researchers used this so-called teacher-student paradigm not only to compress models but also for training state-of-the-art sentiment analysis models  \cite{DBLP:journals/corr/RuderGB17} or multilingual neural machine translation system \cite{multilingual}. \\

What we propose is to build a multi-domain student educated by multiple teachers trained to be experts in a specific domain using state-of-the-art single domain adaptation methods. Our method outperforms typical benchmarks in the domain while being as scalable and fast as a generic translation model.
\\

Our main contribution can be separated in three parts:
\renewcommand\labelitemi{\tiny$\bullet$}
\begin{itemize}
    \item Extending multi-teacher knowledge distillation training framework to multi-domain translation task
    \item Using a pretrained generic student and dynamic data selection while finetuning the teachers
    \item Generalizing the approach for 2, 3 and 4 domains
\end{itemize}

\section{Background} 

\subsection{Neural Machine Translation}
Neural Machine Translation is the field studying the use of neural networks to design automatic ways to perform translation from a language to another.

In our work, we will use the well known Seq2Seq architecture and especially the recent advances in sequence-to-sequence modeling, namely the Transformer that will be presented in Section 2.2.

The Seq2Seq architecture relies on an encoder-decoder mechanism. The very first models of such architecture \cite{DBLP:journals/corr/SutskeverVL14,cho-etal-2014-learning} where based on the following idea: the encoder receives an input sequence which it encodes into a hidden state vector exploited by the decoder to produce the output. In neural machine translation, a source sentence is encoded and then decoded to produce a valid translation. 

\subsubsection{Encoder}
Given an input sequence $\mathbf{x} =  (x_1, x_2, ..., x_{L_x})$ and an output sequence $\mathbf{y} =  (y_1, y_2, ..., y_{L_y})$, the hidden states are computed as following :

\begin{equation}
    h_t^{enc} = f^{enc}(W_{hh}^{enc}h_{t-1}^{enc} + W_{hx}^{enc}x_{t} + b_h^{enc})
\end{equation}

where $t$ $\in$ $1,...,L_x$, and $f^{enc}$ is the encoding function that depends on the chosen architecture. 

\subsubsection{Decoder}
On the decoder side, the initialization of the hidden states may be done using the encoder final hidden state, hence $h_0^{dec} = h_{L_x}^{enc}$, or randomly. Then the decoder's hidden states computation is as follows:

\noindent
If teacher forcing \cite{NIPS2016_6099} is enabled, i.e feeding the network with the ground truth:
\begin{equation}
    h_t^{dec} = g^{dec}(W_{hh}^{dec}h_{t-1}^{dec} + W_{hx}^{dec}y_{t-1} + b_h^{dec})
\end{equation}
\begin{equation}
    h_t^{dec} = g^{dec}(h_{t-1}^{dec}, y_{t-1})
\end{equation}

\noindent
If teacher forcing is disabled, i.e feeding the network with the predicted value at the previous step:

\begin{equation}
    h_t^{dec} = g^{dec}(W_{hh}^{dec}h_{t-1}^{dec} + W_{hx}^{dec}y_{t-1}^{pred} + b_h^{dec})
\end{equation}

\noindent
Eventually the prediction is :
\begin{equation}
    y_t^{pred} = z^{dec}(W_{hy}^{dec}h_{t}^{dec} + b_y^{dec})
\end{equation}
where t $\in$ 1,...,$L_y$.  \\

\subsection{Recurrent Neural Network and Attention Mechanism}

\subsubsection{Recurrent Neural Networks}

Presented computations for encoder and decoder stands for the simplest and the most standard form of recurrent networks. Recurrent Neural Networks, are now a standard architecture that allows dealing with sequential data, as previous output are fed as input to the following computations. It has been improved in order to avoid vanishing and exploding gradient problem through the development of adapted architectures such as LSTM or GRU \cite{hochreiter1997long,cho-etal-2014-learning}.

\subsubsection{Attention Mechanism}

Later, encoding into a single vector a sentence was found limiting, especially for long sentence. \cite{bahdanau2014neural, DBLP:journals/corr/LuongPM15} introduces Attention based Sequence to Sequence networks allowing decoder to consider not only the encoder's last hidden state but all the encoder's hidden states. Attention Networks, are now at the basis of most state-of-the-art neural machine translation architectures. Attention mechanism, makes use of a context vector, that is a weighted average of the encoder's hidden states. Those attention weights are computed using a score that depends on current or previous hidden state and that is learned during the training step. 

\subsection{The Transformer Network}
Until \cite{NIPS2017_7181}, Standard state-of-the-art Seq2Seq models were all based on recurrent neural networks or convolutional neural networks given the sequential nature of the data. However, handling sequences element by element sequentially is an obstacle for parallelization. 

The novelty of the Transformer architecture is its replacement of sequential computations by attentions and positional encoding to keep track of the element's position in the sequence. This enabled to accelerate the training time and to reduce the complexity of computing dependency between elements independently of their position, while recurrent models were obliged to pass through all the intermediate elements.

Transformer is now the leading architecture that helped getting state-of-the-art results in most neural machine translation and natural language processing tasks.

\subsection{Transfer learning}

Transfer learning is a key topic in natural language processing as it led to state-of-art results \cite{DBLP:journals/corr/abs-1907-11692,DBLP:journals/corr/abs-1810-04805, DBLP:journals/corr/abs-1801-06146}.

It relies on the assumption that pre-training neural networks on a large set of data in a task $\mathcal{A}$ will help initialize a network trained on a second task $\mathcal{B}$ where data is scarce. 

The insufficiency of in-domain data has led researchers to train well performing generic models before adapting them on the target domain using in-domain data.


\subsection{Knowledge distillation}
In machine learning, we accept the idea that the objective function is made to reflect the interest of the user as closely as possible. However, all the algorithms tend to minimize the cost function on the training data while the real interest is to generalize well on new data. It is clearly better to train models to generalize better but this requires knowing how to do so. 

When we are distilling the knowledge from a large model to a small model or from a specialist to a generalist, we can train the student to generalize in the same way than the teacher. In general, the teacher is well suited to transfer this kind of information as it is a cumbersome model. 


The objective of knowledge distillation is to fill the gap between the interest of the user, which is good generalization on unseen data, and the cost function used during training. One way to transfer the generalization ability of the cumbersome model is to use class probabilities as soft targets. Instead of trying to match the ground truth labels, we will perform optimization on the softened targets provided by the teacher. 

In other terms, in classification, the negative log-likelihood cost function will be replaced by the Kullback-Leibler divergence between the teacher's distribution and the student distribution.

Traditionally in neural machine translation, the loss function for one sentence is:
\begin{equation}
    l = -\sum_{j = 1}^{sent\_length}\sum_{k = 1}^{|V|}\mathds{1}[y_j=k]\\ log(p(y_j=k|x,y_{<j}))
\end{equation}
where $x$ is the source sentence and $y=(y_1,y_2,\ldots,y_{L_y})$ is the target sentence.

Replacing the cost function by the KL-divergence would result in:
\begin{multline}
    l = -\sum_{j = 1}^{sent\_length}\sum_{k = 1}^{|V|}p^T(y_j=k|x,y_{<j}) \\ log(p^S(y_j=k|x,y_{<j}))
\end{multline}

where $p^T$ and $p^S$ are the probability distributions of the teacher and the student respectively.

This is considered as the simplest form of distillation that can work without having true labels for the transfer set, which can be the teacher's training set. When the correct labels are known for this transfer set or a subset of it, we can incorporate this information to make use of the added information and train the model to produce the correct labels. 

The combination method used in our case result in this cost function which is a weighted sum between the traditional negative log-likelihood and the Kullback-Leibler divergence. 

\begin{multline}
    l = -\sum_{j = 1}^{sent\_length}\sum_{k = 1}^{|V|} ((1-\lambda)\mathds{1}[y_j=k] + \lambda p^T(y_j = k|x,y_{<j})) \\ log(p^S(y_j = k|x,y_{<j}))
\end{multline}

where $\lambda$ is a hyper-parameter of the method quantifying the softening of the targets. If $\lambda$ = 0, we get the cross-entropy loss and if $\lambda$ = 1, we get the simplest form of distillation described above where the ground truth labels are not exploited. \\

\noindent
\textbf{Knowledge distillation as a compression technique}. Knowledge distillation enables to transfer knowledge between a teacher and a student without any size constraint as only class probabilities are used in the cost function. The class probabilities compatibility depend only on the vocabulary which means that the teacher and the student must share the same vocabulary and not the size (number of parameters of both models). 

Therefore, a cumbersome model having hundreds of millions of parameters can educate a simpler student having tens of millions of parameters and producing a similar prediction quality. 


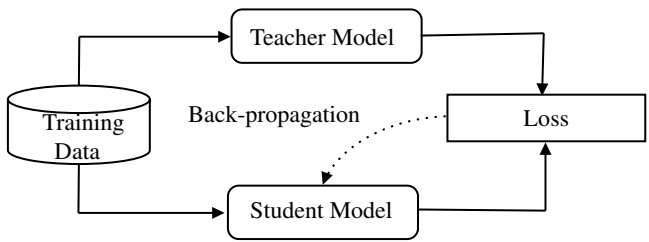
\begin{figure}
\tikzset{every picture/.style={line width=0.75pt}} 

\begin{tikzpicture}[x=0.75pt,y=0.75pt,yscale=-0.7,xscale=0.7]
\draw   (211.2,91.58) -- (211.2,128.22) .. controls (211.2,133.67) and (188.55,138.08) .. (160.6,138.08) .. controls (132.66,138.08) and (110,133.67) .. (110,128.22) -- (110,91.58)(211.2,91.58) .. controls (211.2,97.03) and (188.55,101.45) .. (160.6,101.45) .. controls (132.66,101.45) and (110,97.03) .. (110,91.58) .. controls (110,86.14) and (132.66,81.72) .. (160.6,81.72) .. controls (188.55,81.72) and (211.2,86.14) .. (211.2,91.58) -- cycle ;
\draw   (423.43,88.25) -- (564.5,88.25) -- (564.5,121.75) -- (423.43,121.75) -- cycle ;
\draw   (269.47,34.67) .. controls (269.47,30.42) and (272.91,26.99) .. (277.15,26.99) -- (397.35,26.99) .. controls (401.59,26.99) and (405.03,30.42) .. (405.03,34.67) -- (405.03,57.7) .. controls (405.03,61.94) and (401.59,65.38) .. (397.35,65.38) -- (277.15,65.38) .. controls (272.91,65.38) and (269.47,61.94) .. (269.47,57.7) -- cycle ;
\draw   (266.9,161.37) .. controls (266.9,157.08) and (270.37,153.61) .. (274.66,153.61) -- (394.69,153.61) .. controls (398.98,153.61) and (402.45,157.08) .. (402.45,161.37) -- (402.45,184.24) .. controls (402.45,188.53) and (398.98,192) .. (394.69,192) -- (274.66,192) .. controls (270.37,192) and (266.9,188.53) .. (266.9,184.24) -- cycle ;
\draw    (160.91,46.59) -- (160.6,81.72) ;

\draw    (160.91,46.59) -- (266.86,46.59) ;
\draw [shift={(268.86,46.59)}, rotate = 180] [fill={rgb, 255:red, 0; green, 0; blue, 0 }  ][line width=0.75]  [draw opacity=0] (8.93,-4.29) -- (0,0) -- (8.93,4.29) -- cycle    ;

\draw    (492.12,44.14) -- (490.95,87.07) ;
\draw [shift={(490.9,89.07)}, rotate = 271.56] [fill={rgb, 255:red, 0; green, 0; blue, 0 }  ][line width=0.75]  [draw opacity=0] (8.93,-4.29) -- (0,0) -- (8.93,4.29) -- cycle    ;

\draw    (404.41,45.78) -- (492.12,44.14) ;

\draw    (160.3,173.21) -- (262.57,173.21) ;
\draw [shift={(264.57,173.21)}, rotate = 180] [fill={rgb, 255:red, 0; green, 0; blue, 0 }  ][line width=0.75]  [draw opacity=0] (8.93,-4.29) -- (0,0) -- (8.93,4.29) -- cycle    ;

\draw    (160.6,138.08) -- (160.3,173.21) ;

\draw    (493.35,171.58) -- (493.35,124.56) ;
\draw [shift={(493.35,122.56)}, rotate = 450] [fill={rgb, 255:red, 0; green, 0; blue, 0 }  ][line width=0.75]  [draw opacity=0] (8.93,-4.29) -- (0,0) -- (8.93,4.29) -- cycle    ;

\draw    (402.57,170.76) -- (493.35,171.58) ;

\draw  [dash pattern={on 0.84pt off 2.51pt}]  (422.2,103.78) .. controls (385.25,104.98) and (347.1,122.85) .. (335.2,151.25) ;
\draw [shift={(334.5,153)}, rotate = 290.58] [fill={rgb, 255:red, 0; green, 0; blue, 0 }  ][line width=0.75]  [draw opacity=0] (8.93,-4.29) -- (0,0) -- (8.93,4.29) -- cycle    ;

\draw (163.98,118.48) node  [align=left] {{\fontfamily{ptm}\selectfont Training }\\{\fontfamily{ptm}\selectfont  \ \ Data}};
\draw (493.96,105) node  [align=left] {{\fontfamily{ptm}\selectfont Loss}};
\draw (334.18,47) node  [align=left] {{\small {\fontfamily{ptm}\selectfont Teacher Model}}};
\draw (333.57,171.17) node  [align=left] {{\small {\fontfamily{ptm}\selectfont Student Model}}};
\draw (299.92,103.68) node  [align=left] {{\small Back-propagation}};

\end{tikzpicture}
\caption{How compression works through distillation?}
\end{figure}

In \cite{distillation}, Geoffrey Hinton et al. investigated the effects of ensembling Deep Neural Network acoustic models that are used in Automatic Speech Recognition. They show it is possible to distill the knowledge of an ensemble of 10 models to a single model achieving similar Word Error Rate (WER) having ten times less parameters.

\section{Related Work} 
Domain adaptation for machine translation has been largely explored. 

In a survey summarizing the different developed methods \cite{DBLP:journals/corr/abs-1806-00258}, it is observed that most of the multi-domain adaptation techniques are based on new architectures with additional domain classifiers which reduces their scalability compared to the standard model, and only a few try to leverage data or the training objective. 



The majority of the articles study adaptation \textbf{on a single domain}. Scalable methods are essentially based on finetuning using either in-domain data only or a mix of out-of-domain and in-domain data in order to reduce overfitting \cite{DBLP:journals/corr/ServanCS16,luong-etal-2015,DBLP:journals/corr/FreitagA16,van-der-wees-etal-2017-dynamic}. Other methods try to leverage the loss function by introducing a weighting strategy \cite{wang-etal-2017-instance}. 

While multiple articles tackle the \textbf{multi-domain adaption problem} with promising results, most of them rely either on ensembling \cite{DBLP:journals/corr/FreitagA16}, a priori domain clustering in order to add domain tags \cite{DBLP:journals/corr/abs-1805-02282} or introducing a new domain specific gating vector \cite{zeng-etal-2018-multi}. So far, the best technique that does not add more complexity or prior classification, either using supervised or unsupervised methods, is based on finetuning on the concatenation of all in-domain data \cite{DBLP:journals/corr/abs-1708-08712} \cite{DBLP:journals/corr/abs-1806-00258}. 

\textbf{Knowledge distillation} is a well-known neural network compression technique, especially for machine translation \cite{distillation,kim-rush-2016-sequence,DBLP:journals/corr/FreitagAS17}. Recently,  \cite{DBLP:journals/corr/RuderGB17} showed that relying on multiple experts teachers allowed to improve the performances of sentiment analysis on unseen domains. In \cite{multilingual}, the authors applied the multiple teachers framework in knowledge distillation in order to build state-of-the-art multilingual machine translation system. This work applied to multilingual machine translation is the closest to the one that we propose: it uses a similar methodology to the one we explored but it fulfills a different goal. Highly motivated by those results, our goal was to apply the multiple teachers knowledge distillation framework in order to improve the performances on multi-domain neural machine translation. So far, to our knowledge, the only usage of knowledge distillation in the context of neural machine translation adaptation is limiting the performance degradation on the out-of-domain data \cite{dakwale2017fine}.

\section{Approach} 

Neural Machine Translation models are often trained on large open-source data coming from institutions such as the European parliament, the European commission, movies subtitles, Wikipedia pages, etc. These corpora offer us the necessary amount of data to train neural networks having hundreds of millions of parameters. However these models are often not well performing on specialized datasets and therefore transfer learning may be used.

In our approach, we use \textit{Dynamic Data Selection}, a transfer learning technique, to train models on specialized datasets that will play a teacher role when building a multi-domain student through \textit{a multi-task knowledge distillation strategy}. 

\subsection{Knowledge distillation as a multi-task technique}
Initially, we presented the distillation strategy as a way to perform compression on neural networks. This has been proved to reduce the complexity while preserving the performance of the cumbersome network. In our work, we used knowledge distillation to produce a multi-domain student that may or may not be of the same size than the teachers. 


\hspace{0.5pt}

\tikzset{every picture/.style={line width=0.75pt}} 
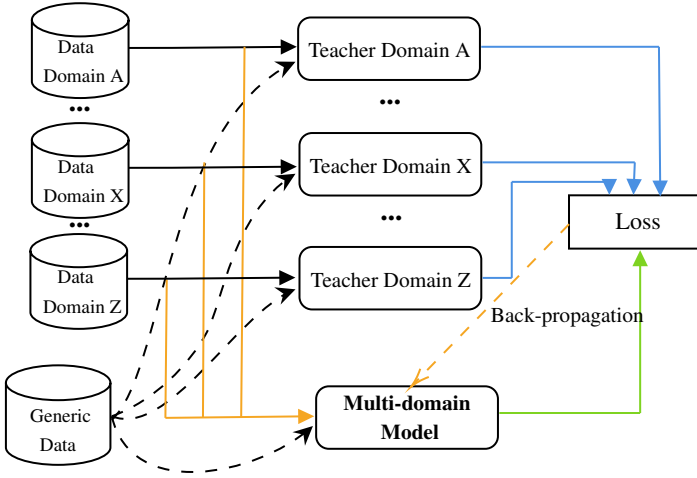
\begin{figure}
\begin{tikzpicture}[x=0.75pt,y=0.75pt,yscale=-0.9,xscale=0.9]

\draw   (122.52,31.28) -- (122.52,65.73) .. controls (122.52,70.85) and (110.52,75) .. (95.71,75) .. controls (80.9,75) and (68.9,70.85) .. (68.9,65.73) -- (68.9,31.28)(122.52,31.28) .. controls (122.52,36.4) and (110.52,40.55) .. (95.71,40.55) .. controls (80.9,40.55) and (68.9,36.4) .. (68.9,31.28) .. controls (68.9,26.15) and (80.9,22) .. (95.71,22) .. controls (110.52,22) and (122.52,26.15) .. (122.52,31.28) -- cycle ;
\draw   (366.81,129.36) -- (440.5,129.36) -- (440.5,159.15) -- (366.81,159.15) -- cycle ;
\draw   (216.58,36.95) .. controls (216.58,33.1) and (219.71,29.97) .. (223.57,29.97) -- (310.43,29.97) .. controls (314.29,29.97) and (317.41,33.1) .. (317.41,36.95) -- (317.41,57.91) .. controls (317.41,61.77) and (314.29,64.89) .. (310.43,64.89) -- (223.57,64.89) .. controls (219.71,64.89) and (216.58,61.77) .. (216.58,57.91) -- cycle ;
\draw   (226.57,243.16) .. controls (226.57,239.27) and (229.73,236.11) .. (233.63,236.11) -- (320.34,236.11) .. controls (324.24,236.11) and (327.4,239.27) .. (327.4,243.16) -- (327.4,263.97) .. controls (327.4,267.87) and (324.24,271.03) .. (320.34,271.03) -- (233.63,271.03) .. controls (229.73,271.03) and (226.57,267.87) .. (226.57,263.97) -- cycle ;
\draw   (217.33,101.54) .. controls (217.33,97.68) and (220.45,94.55) .. (224.31,94.55) -- (311.17,94.55) .. controls (315.03,94.55) and (318.16,97.68) .. (318.16,101.54) -- (318.16,122.49) .. controls (318.16,126.35) and (315.03,129.48) .. (311.17,129.48) -- (224.31,129.48) .. controls (220.45,129.48) and (217.33,126.35) .. (217.33,122.49) -- cycle ;
\draw   (217.33,165.21) .. controls (217.33,161.35) and (220.45,158.22) .. (224.31,158.22) -- (311.17,158.22) .. controls (315.03,158.22) and (318.16,161.35) .. (318.16,165.21) -- (318.16,186.16) .. controls (318.16,190.02) and (315.03,193.15) .. (311.17,193.15) -- (224.31,193.15) .. controls (220.45,193.15) and (217.33,190.02) .. (217.33,186.16) -- cycle ;
\draw    (122.52,47) -- (146.67,46.96) -- (213.86,46.37) ;
\draw [shift={(215.86,46.35)}, rotate = 539.49] [fill={rgb, 255:red, 0; green, 0; blue, 0 }  ][line width=0.75]  [draw opacity=0] (8.93,-4.29) -- (0,0) -- (8.93,4.29) -- cycle    ;

\draw   (112.47,234.12) -- (112.47,270.27) .. controls (112.47,275.64) and (99.49,280) .. (83.48,280) .. controls (67.48,280) and (54.5,275.64) .. (54.5,270.27) -- (54.5,234.12)(112.47,234.12) .. controls (112.47,239.49) and (99.49,243.85) .. (83.48,243.85) .. controls (67.48,243.85) and (54.5,239.49) .. (54.5,234.12) .. controls (54.5,228.74) and (67.48,224.38) .. (83.48,224.38) .. controls (99.49,224.38) and (112.47,228.74) .. (112.47,234.12) -- cycle ;
\draw [color={rgb, 255:red, 74; green, 144; blue, 226 }  ,draw opacity=1 ]   (318.14,46.35) -- (416.7,46.35) ;

\draw [color={rgb, 255:red, 74; green, 144; blue, 226 }  ,draw opacity=1 ]   (318.88,110.94) -- (403.31,110.94) ;

\draw [color={rgb, 255:red, 74; green, 144; blue, 226 }  ,draw opacity=1 ]   (318.88,175.52) -- (335.06,175.52) ;

\draw [color={rgb, 255:red, 74; green, 144; blue, 226 }  ,draw opacity=1 ]   (335.06,121.85) -- (335.06,175.52) ;

\draw [color={rgb, 255:red, 74; green, 144; blue, 226 }  ,draw opacity=1 ]   (335.06,121.85) -- (389.17,121.85) ;

\draw [color={rgb, 255:red, 74; green, 144; blue, 226 }  ,draw opacity=1 ]   (416.7,46.35) -- (416.95,74.53) -- (417.42,128.04) ;
\draw [shift={(417.44,130.04)}, rotate = 269.49] [fill={rgb, 255:red, 74; green, 144; blue, 226 }  ,fill opacity=1 ][line width=0.75]  [draw opacity=0] (8.93,-4.29) -- (0,0) -- (8.93,4.29) -- cycle    ;

\draw [color={rgb, 255:red, 74; green, 144; blue, 226 }  ,draw opacity=1 ]   (403.31,110.94) -- (402.65,127.13) ;
\draw [shift={(402.56,129.13)}, rotate = 272.34000000000003] [fill={rgb, 255:red, 74; green, 144; blue, 226 }  ,fill opacity=1 ][line width=0.75]  [draw opacity=0] (8.93,-4.29) -- (0,0) -- (8.93,4.29) -- cycle    ;

\draw [color={rgb, 255:red, 74; green, 144; blue, 226 }  ,draw opacity=1 ]   (389.17,121.85) -- (389.17,127.13) ;
\draw [shift={(389.17,129.13)}, rotate = 270] [fill={rgb, 255:red, 74; green, 144; blue, 226 }  ,fill opacity=1 ][line width=0.75]  [draw opacity=0] (8.93,-4.29) -- (0,0) -- (8.93,4.29) -- cycle    ;

\draw [color={rgb, 255:red, 126; green, 211; blue, 33 }  ,draw opacity=1 ]   (406.28,251.02) -- (406.28,162.06) ;
\draw [shift={(406.28,160.06)}, rotate = 450] [fill={rgb, 255:red, 126; green, 211; blue, 33 }  ,fill opacity=1 ][line width=0.75]  [draw opacity=0] (8.93,-4.29) -- (0,0) -- (8.93,4.29) -- cycle    ;

\draw [color={rgb, 255:red, 126; green, 211; blue, 33 }  ,draw opacity=1 ]   (328.18,251.02) -- (406.28,251.02) ;

\draw [color={rgb, 255:red, 245; green, 166; blue, 35 }  ,draw opacity=1 ] [dash pattern={on 4.5pt off 4.5pt}]  (366.86,145.5) -- (305.98,208.59) -- (280.47,235.02) ;
\draw [shift={(279.09,236.46)}, rotate = 313.98] [color={rgb, 255:red, 245; green, 166; blue, 35 }  ,draw opacity=1 ][line width=0.75]    (10.93,-3.29) .. controls (6.95,-1.4) and (3.31,-0.3) .. (0,0) .. controls (3.31,0.3) and (6.95,1.4) .. (10.93,3.29)   ;

\draw  [dash pattern={on 4.5pt off 4.5pt}]  (113.21,254.84) .. controls (142.81,227.68) and (142.96,89.77) .. (214.04,55.23) ;
\draw [shift={(215.12,54.72)}, rotate = 514.99] [fill={rgb, 255:red, 0; green, 0; blue, 0 }  ][line width=0.75]  [draw opacity=0] (10.72,-5.15) -- (0,0) -- (10.72,5.15) -- (7.12,0) -- cycle    ;

\draw  [dash pattern={on 4.5pt off 4.5pt}]  (113.21,252.84) .. controls (137.01,244.24) and (156.35,223.73) .. (165.69,206.36) .. controls (174.88,189.26) and (180,139.74) .. (213.56,118.27) ;
\draw [shift={(215.12,117.3)}, rotate = 509.1] [fill={rgb, 255:red, 0; green, 0; blue, 0 }  ][line width=0.75]  [draw opacity=0] (10.72,-5.15) -- (0,0) -- (10.72,5.15) -- (7.12,0) -- cycle    ;

\draw  [dash pattern={on 4.5pt off 4.5pt}]  (113.21,252.84) .. controls (144.71,265.38) and (181.99,195.85) .. (213.67,182.46) ;
\draw [shift={(215.12,181.89)}, rotate = 519.71] [fill={rgb, 255:red, 0; green, 0; blue, 0 }  ][line width=0.75]  [draw opacity=0] (10.72,-5.15) -- (0,0) -- (10.72,5.15) -- (7.12,0) -- cycle    ;

\draw  [dash pattern={on 4.5pt off 4.5pt}]  (113.21,252.84) .. controls (124.2,300.32) and (192.93,286.03) .. (223.42,259.51) ;
\draw [shift={(224.79,258.29)}, rotate = 497.47] [fill={rgb, 255:red, 0; green, 0; blue, 0 }  ][line width=0.75]  [draw opacity=0] (10.72,-5.15) -- (0,0) -- (10.72,5.15) -- (7.12,0) -- cycle    ;

\draw [color={rgb, 255:red, 245; green, 166; blue, 35 }  ,draw opacity=1 ]   (142.96,253.75) -- (223.53,252.86) ;
\draw [shift={(225.53,252.84)}, rotate = 539.37] [fill={rgb, 255:red, 245; green, 166; blue, 35 }  ,fill opacity=1 ][line width=0.75]  [draw opacity=0] (8.93,-4.29) -- (0,0) -- (8.93,4.29) -- cycle    ;

\draw [color={rgb, 255:red, 245; green, 166; blue, 35 }  ,draw opacity=1 ]   (186.85,46.35) -- (184.99,253.29) ;

\draw [color={rgb, 255:red, 245; green, 166; blue, 35 }  ,draw opacity=1 ]   (143.71,176.43) -- (142.96,253.75) ;

\draw [color={rgb, 255:red, 245; green, 166; blue, 35 }  ,draw opacity=1 ]   (164.53,110.94) -- (163.79,253.75) ;

\draw   (122.52,98.28) -- (122.52,132.73) .. controls (122.52,137.85) and (110.52,142) .. (95.71,142) .. controls (80.9,142) and (68.9,137.85) .. (68.9,132.73) -- (68.9,98.28)(122.52,98.28) .. controls (122.52,103.4) and (110.52,107.55) .. (95.71,107.55) .. controls (80.9,107.55) and (68.9,103.4) .. (68.9,98.28) .. controls (68.9,93.15) and (80.9,89) .. (95.71,89) .. controls (110.52,89) and (122.52,93.15) .. (122.52,98.28) -- cycle ;
\draw    (122.52,114) -- (146.67,113.96) -- (213.86,113.37) ;
\draw [shift={(215.86,113.35)}, rotate = 539.49] [fill={rgb, 255:red, 0; green, 0; blue, 0 }  ][line width=0.75]  [draw opacity=0] (8.93,-4.29) -- (0,0) -- (8.93,4.29) -- cycle    ;

\draw   (121.72,160.28) -- (121.72,194.73) .. controls (121.72,199.85) and (109.72,204) .. (94.91,204) .. controls (80.1,204) and (68.1,199.85) .. (68.1,194.73) -- (68.1,160.28)(121.72,160.28) .. controls (121.72,165.4) and (109.72,169.55) .. (94.91,169.55) .. controls (80.1,169.55) and (68.1,165.4) .. (68.1,160.28) .. controls (68.1,155.15) and (80.1,151) .. (94.91,151) .. controls (109.72,151) and (121.72,155.15) .. (121.72,160.28) -- cycle ;
\draw    (121.72,176) -- (145.87,175.96) -- (213.06,175.37) ;
\draw [shift={(215.06,175.35)}, rotate = 539.49] [fill={rgb, 255:red, 0; green, 0; blue, 0 }  ][line width=0.75]  [draw opacity=0] (8.93,-4.29) -- (0,0) -- (8.93,4.29) -- cycle    ;

\draw (97.7,54.29) node  [align=left] {{\fontfamily{ptm}\selectfont {\scriptsize  \ \ \ Data }}\\{\fontfamily{ptm}\selectfont {\scriptsize Domain A}}};
\draw (405.14,143.68) node  [align=left] {{\fontfamily{ptm}\selectfont Loss}};
\draw (267.28,48.17) node  [align=left] {{\footnotesize {\fontfamily{ptm}\selectfont Teacher Domain A}}};
\draw (276.98,253.57) node  [align=left] {{\footnotesize {\fontfamily{ptm}\selectfont \textbf{Multi-domain }}}\\{\footnotesize {\fontfamily{ptm}\selectfont \textbf{ \ \ \ \ \ \ Model}}}};
\draw (268.03,112.76) node  [align=left] {{\footnotesize {\fontfamily{ptm}\selectfont Teacher Domain X}}};
\draw (268.03,176.43) node  [align=left] {{\footnotesize {\fontfamily{ptm}\selectfont Teacher Domain Z}}};
\draw (82.21,260.5) node  [align=left] {{\fontfamily{ptm}\selectfont {\scriptsize  \ \ Generic }}\\{\fontfamily{ptm}\selectfont {\scriptsize  \ \ \ \ Data}}};
\draw (365.88,197.35) node  [align=left] {{\footnotesize {\fontfamily{ptm}\selectfont Back-propagation}}};
\draw (267.56,77) node  [align=left] {\textbf{{\large ...}}};
\draw (268.41,142) node  [align=left] {\textbf{{\large ...}}};
\draw (97.7,121.29) node  [align=left] {{\fontfamily{ptm}\selectfont {\scriptsize  \ \ \ Data }}\\{\fontfamily{ptm}\selectfont {\scriptsize Domain X}}};
\draw (96.9,183.29) node  [align=left] {{\fontfamily{ptm}\selectfont {\scriptsize  \ \ \ Data }}\\{\fontfamily{ptm}\selectfont {\scriptsize Domain Z}}};
\draw (95.56,81) node  [align=left] {\textbf{{\large ...}}};
\draw (95.41,146) node  [align=left] {\textbf{{\large ...}}};
\end{tikzpicture}
\caption{Training a multi-domain student using multiple single domain teachers}
\end{figure}

 This strategy enables us to use a single model to translate sentences from different domains instead of storing different expert models, filtering the data during inference and using the corresponding teacher. It is time and memory efficient and particularly convenient in a production environment.

The use of specialists that are trained on different domains has some resemblance to mixture of experts which learns how to assign each example to the most likely expert through probability computation. During training, two learning phases happen simultaneously, the experts are learning how to be accurate on the examples assigned to them and the gating network is learning how to assign examples to the corresponding experts.

The major difference between the two methods is parallelization. It is easier to train at the same time different teachers on their corresponding domain than to parallelize mixtures of experts.\\

\subsection{Dynamic Data Selection}

Here we present a transfer learning technique called \textit{Dynamic data selection} developed by Marlies van der Wees et al. in \cite{van-der-wees-etal-2017-dynamic} used to build the teachers during the distillation process.

The idea is to select sentences from generic data that are similar to the in-domain corpus and at the same time dissimilar to the generic corpus. The metric used for the ranking is based on the cross-entropy.

Let I be the in-domain data, $G$ the generic data. We assume that $G$ contains a subset $G_{I}$ that has the same distribution than I. \\

The objective of the method is to select sentences $s$ from $G$ that are most probably belonging to $G_{I}$, i.e. maximizing $P(G_{I} | s, G)$.

\begin{equation}
    P(G_{I} | s,G) = \frac{P(s|I) P(G_{I}|G)}{P(s|G)}
\end{equation}

This quantity, log-transformed, is close to : $H_{I}(s) - H_{G}(s)$.\\

The entropy $H_{A}(b)$ is defined as: $H_{A}(b) = E_{x \sim P}(-logQ(b))$, P being the language model computed on the corpus A, Q the estimated language model and b the sentence of interest.\\

In order to perform data selection on a bi-text corpus, a ranking metric is introduced:

\begin{dmath}
    CED_{s} = [H_{I}(s_{source}) - H_{G}(s_{source})] + [H_{I}(s_{target}) - H_{G}(s_{target})]
\end{dmath}

For example, when calculating $H_{I}(s_{source})$, we build a language model on the in-domain corpus then we use the \textbf{score} function to rank sentences. The output is the log-transformed product of each word's probability. 

During gradual fine-tuning, the selection size $n$ is a function of the epoch $i$ :
\begin{equation}
    n(i) = \alpha  \vert G \vert \beta^{\frac{i-1}{\nu}}
\end{equation}
where $\alpha$ is the relative start size i.e. the fraction of the out-of-domain data for the first selection, $G$ is the size of the latter, 0 $\leq \beta \leq$ 1 is the retention rate i.e. the fraction to be kept at each selection, $i$ the epoch, $\nu$ the number of epochs where the selected subset doesn't change. \\

This finetuning approach has been proven to provide state-of-the-art result in the context of domain specialization (especially when the set of in-domain parallel sentences is small) while significantly reducing the training time. This method have been used in order to finetune our expert teachers on single domains.

\subsection{Distillation process}

The training that we propose for distillation process is described in \textbf{Algorithm 1}.

\begin{algorithm}
\caption{Multi-domain student training pipeline}
    \textbf{Input:} Generic dataset $\mathcal{G}$, $n$ different specialiazied datasets $(\mathcal{D}_i)_{i\in \{1..n\}}$
    
    \textbf{Output:} Multi-domain student model $S_{multi}$ trained using knowledge distillation
    
    \textbf{Initialization:} Transformer model $M_{gen}$ initialized with random weights. 
    
    Empty array $\mathcal{P}_{T}$ of size (T,L,$\vert V \vert$) where T is the sentence length, L the training set size and $\vert V \vert$ the vocabulary size.

\textit{\textbf{1. Train generic model:}}
\begin{algorithmic}
\WHILE{$M_{gen}$ not converging}{}
\STATE \textit{Train} $M_{gen}$ with batches from $\mathcal{G}$
\ENDWHILE
\end{algorithmic}    
 
\noindent
\textit{\textbf{2. Finetune teachers:}}
\begin{algorithmic}
\FOR{$i$ in  $\{1..n\}$}{}
 \STATE \textit{Initialize} $\mathcal{T}_{i}$ = $M_{gen}$ 
\STATE \textit{Finetune} $\mathcal{T}_{i}$ using \textit{dynamic data selection} on specialized dataset $\mathcal{D}_i$ and generic data $\mathcal{G}$.
\ENDFOR
\end{algorithmic}

\noindent
\textit{\textbf{3. Extend $\mathcal{P}_{T}$:}}
   
\begin{algorithmic}
\FOR{$i$ in  $\{1..n\}$}{}
\FOR{$d$ in  $\mathcal{D}_i$}{}
\STATE \textit{Compute} $\mathcal{P}_{T_{i}}(d_{src})$, the probability distribution resulting from $\mathcal{T}_{i}$'s translation of $d_{src}$.
\STATE \textit{Append} $\mathcal{P}_{T_{i}}(d_{src})$ to $\mathcal{P}_{T}$ 
\ENDFOR
\ENDFOR
\end{algorithmic}

\noindent
\textit{\textbf{4. Train student model:}}

Let $\mathcal{G}_{sub}$ be a randomly selected subset from the generic dataset.
\begin{algorithmic}
\WHILE{$S_{multi}$ not converging}{}

\STATE \textit{Train} $S_{multi}$ with batches from ($\cup_{i\in \{1..n\}} \mathcal{D}_i$) $\cap \ \mathcal{G}_{sub}$  using $\mathcal{P}_{T}$ in the cost function.
\ENDWHILE
\end{algorithmic} 
\end{algorithm}

\subsection{Top-K distillation}

In our experiments, the student model does not match the full distribution of the teacher model but only the top-K output distribution in order to reduce memory cost. Actually, it is expensive in terms of RAM memory to load the full distributions of the different teacher models knowing that we may have a large number of teachers. 

\subsection{Word-Level knowledge distillation}

In Neural Machine Translation, two kind of distillation processes have been explored so far. The first one, word-level is based on training the student to mimick teacher's local word distributions (using the ground truth target sequence) while the second one, sequence-level is based on training the student to mimick the teacher's output after beam search (without using the ground truth target sequence).

While \cite{kim-rush-2016-sequence} have proved sequence level knowledge distillation efficiency for neural network compression. \cite{multilingual} proved that sequence level knowledge distillation when used in the multi-teacher framework results in significantly lower performances than word-level distillation.

Hence, all our experiments are based on word-level knowledge distillation only.

\subsection{Label smoothing impact on distillation}
In \cite{label_smoothing_help}, the authors investigate the consequences of using label smoothing, when training teachers, on the student performance. They've shown that using label smoothing does not necessarily lead to better distillation.

They assume that this shortcoming is linked to the information erasure caused by label smoothing. A visualization technique applied to the penultimate layer representations of image classifiers trained on image datasets such as CIFAR10 enabled to show that using label smoothing results in a loss of information concerning similarities between examples of different classes.

\section{Experiments and Results} 
\subsection{Experimental setup}

\subsubsection{Datasets and preprocessing}

In order to assess our method's efficiency we focused on the English to French multi-domain translation. First we start by building the initial \textbf{generic} model used for the transfer learning tasks. The initial generic model was trained using WMT14 data preprocessed using a standard procedure: 40k operations based BPE joint-tokenization (source and target sentences are sharing the same BPE tokenization), filtering sentences longer than 250 tokens and sentences with a ratio between source sentence and target sentence length higher than 1.5 \cite{DBLP:journals/corr/abs-1808-09381}. 

Domain \textbf{specific} data will focus on 4 domains: Medical, Legal, Software documentation, and religion. Indeed, the experiments will be conducted on open-source data: EMEA\footnote{http://opus.nlpl.eu/EMEA.php} [Medical] (European Medicines Agency documents), JRC\footnote{http://opus.nlpl.eu/JRC-Acquis.php} [Legal] (a collection of legislative text of the European Union),  GNOME  \footnote{http://opus.nlpl.eu/GNOME.php} [Software] and BIBLE \footnote{http://opus.nlpl.eu/bible-uedin.php} [Religion] corpora . Specialized data is processed in the same way as the generic one. All the datasets are splitted in a training and a testing part which is the first 2000 sentences of the corpus.

\subsubsection{Hyperparameters and settings}

During the experiment we trained a Transformer Base network implemented in OpenNMT-py framework \cite{DBLP:journals/corr/KleinKDSR17}, using Adam optimizer ($\beta_1 = 0.9, \beta_2 = 0.98$), label smoothing and dropout equal to $0.1$, with noam decay and an initial learning rate equal to $2$.

The generic model was trained during 130k steps on 4 V100 GPUs, with a token-based batch size of 4096 and gradient accumulation during 2 steps (equivalent to 8 V100 training). Following common postprocessing we averaged model checkpoints during steps 110k, 120k, and 130k in order to get the final model.

In order to build the teachers used during knowledge distillation, this pretrained model was finetuned for 10k steps on each specialized dataset. Checkpoints were saved every 1000 steps and are used to get the best model according to the criterion described in the Results section. 

This process generated four expert teachers for Medical, Legal, Software and Religion datasets.

\subsection{Results}

\subsubsection{Distillation strategies}

In our first experiment, we compare the performance according to three different strategies : no distillation ($\lambda = 0$), pure distillation ($\lambda = 1$), and flexible distillation (0 \textless \ $\lambda < 1$). You can notice that the first strategy corresponds to finetuning.

For the flexible distillation, we use greedy search to find the best value for the parameter  $\lambda$ combining the negative log-likelihood and the Kulback-Leibler divergence. $\lambda = 0.7$ was found to yield the best results.

The performances are evaluated using the BLEU score implemented in \textit{sacrebleu} \cite{post-2018-call}. 

\begin{table}
{\caption{Performance of the teachers, trained without label smoothing, on their corresponding test set}\label{table1}}
\begin{center}
  \begin{tabular}{| c | c | c |}
\hline
\textbf{Teacher} & \textbf{Domain test set} & \textbf{WMT14 test} \\ \hline 
\hline
Teacher Legal &  58.8 &  35.2\\ \hline
Teacher Medical & 58.7 & 33.4\\ \hline
Teacher Software &  37.6 &  30.5\\ \hline
Teacher Religion &  27.8 & 25.4\\ \hline
  \end{tabular}
\end{center}
\end{table}

\begin{table}
{\caption{BLEU scores corresponding to four configurations with 2 to 4 domains using three different distillation strategies}\label{table2}}
\begin{center}
\begin{adjustbox}{width=\columnwidth,center}
  \begin{tabular}{| c |c |c |c| c |c | c|}
    \hline
    \textbf{Configurations} & \textbf{$\lambda$} & \textbf{Legal} & \textbf{Medical} & \textbf{Software} & \textbf{Religion} & \textbf{WMT14}\\ \hline
    \hline
    Initial Generic Model & - & 51.0 &  33.3& 24.6  & 11.5 & 38 \\ \hline
    \begin{tabular}{@{}c@{}}Legal \\ Medical \end{tabular} & \begin{tabular}{@{}c@{}}0 \\ 0.7 \\ 1\end{tabular} & \begin{tabular}{@{}c@{}} 57.4 \\ \textbf{58.4} \\ 58\end{tabular}  & 
       \begin{tabular}{@{}c@{}} 50.0 \\ 53.0 \\ \textbf{54.1}\end{tabular}& - & -  & \begin{tabular}{@{}c@{}} 34.4 \\ \textbf{35.2} \\ 33.7\end{tabular}\\ \hline

    \begin{tabular}{@{}c@{}}Legal - Medical \\ Software \end{tabular} & \begin{tabular}{@{}c@{}}0 \\ 0.7 \\ 1\end{tabular} & \begin{tabular}{@{}c@{}}57.0 \\ \textbf{57.9} \\ 57.6 \end{tabular}  & 
     \begin{tabular}{@{}c@{}}50.0 \\ 52.7 \\ \textbf{53.1} \end{tabular}  & \begin{tabular}{@{}c@{}}\textbf{40.5} \\ 40.2 \\ 39.3 \\\end{tabular} & -  & \begin{tabular}{@{}c@{}} 35.1 \\ \textbf{35.4} \\ 33.6 \end{tabular}\\ \hline

    \begin{tabular}{@{}c@{}}Legal - Medical \\ Software - Religion \end{tabular} & \begin{tabular}{@{}c@{}}0 \\ 0.7 \\ 1\end{tabular} & \begin{tabular}{@{}c@{}}56.4 \\ \textbf{56.7} \\ \textbf{56.7} \end{tabular}  & 
     \begin{tabular}{@{}c@{}}48.8 \\ 51.4 \\ \textbf{52.1}\end{tabular}& \begin{tabular}{@{}c@{}} 37.5 \\ 37.7 \\ \textbf{38.7} \end{tabular} & \begin{tabular}{@{}c@{}}24.4 \\ 27.3 \\ \textbf{27.9} \end{tabular} & \begin{tabular}{@{}c@{}} \textbf{34.9} \\ 34.0 \\ 29.4 \end{tabular}\\ \hline
  \end{tabular}
 \end{adjustbox}
\end{center}
\end{table}

\textit{The scores shown in Table \ref{table2} correspond to the best step for each configuration. The ranking criterion is the average BLEU score over the specialized datasets available in the configuration.}\\

To conduct our experiments, we decided to consider datasets of different difficulties where the most difficult to translate would be in the last configuration. You can see in Table \ref{table1} that Legal and Medical, studied in the first configuration, are the easiest ones to translate (BLEU score of 58.8 and 58.7 respectively) whereas the Software dataset, studied in the configuration that follows, is more challenging (BLEU score of 37.6) and eventually the Religion dataset introduced in the last configuration corresponds to the lowest performance (BLEU score of 27.8). 

The first experiment showed that the pure distillation strategy ($\lambda = 1$) yields the best results on the specialized datasets. However, we see that the BLEU score on the generic dataset WMT14 decreased substantially. In fact, in the last configuration when using $\lambda = 1$, a decrease of 4.6 points in BLEU score is observed on the WMT14 test set compared to the model using $\lambda = 0.7$.

As we want to avoid overfitting on the specialized datasets, we choose the model using $\lambda = 0.7$ to conduct the next experiment.

\subsubsection{Impact of label smoothing}

In the following experiment we compare three models: the student using teachers trained with hard targets, the student using teachers trained with soft targets and, finetuned model (literature optimal benchmark \cite{DBLP:journals/corr/abs-1708-08712}) and the ensembled model. 

The results are summarized similarly to the previous experiment where for each configuration the BLEU scores are computed on the corresponding test sets using the following models: Hard student (\textbf{HS}), Soft student (\textbf{SS}), Ensembling (\textbf{Ensemble}), and Finetuned (\textbf{F}). Ensembling relies on ensembling all the involved single domain teachers.

\begin{figure}[H]
\begin{subfigure}{.25\textwidth}
  \centering
  \includegraphics[width=\linewidth]{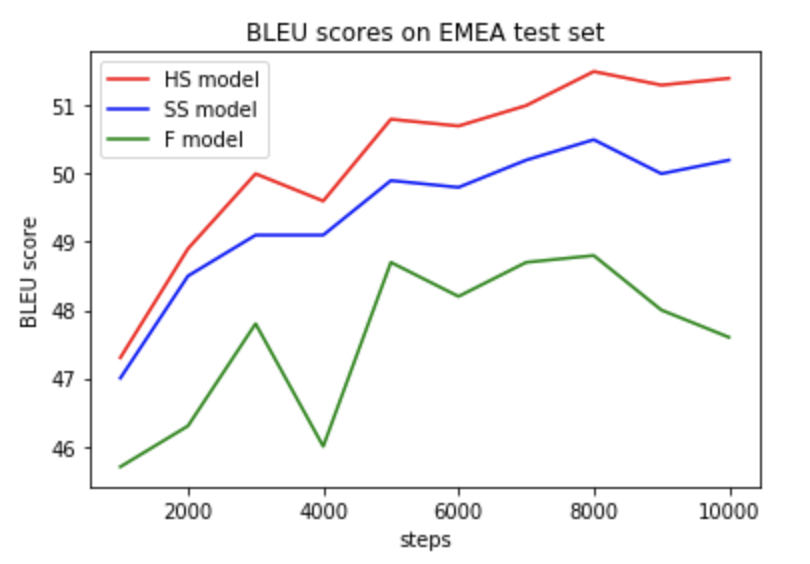}
  \label{fig:sfig1}
\end{subfigure}%
\begin{subfigure}{.25\textwidth}
  \centering
  \includegraphics[width=\linewidth]{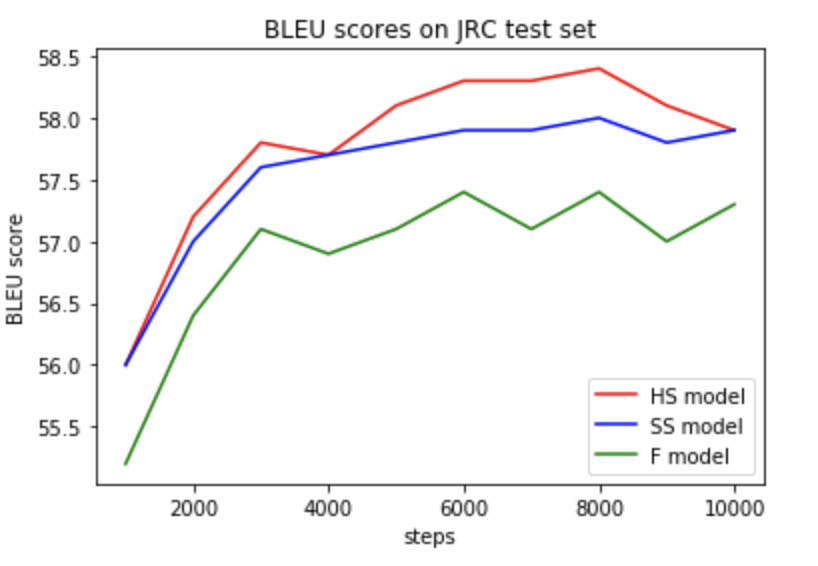}
  \label{fig:sfig2}
\end{subfigure} %
\begin{subfigure}{.25\textwidth}
  \centering
  \includegraphics[width=\linewidth]{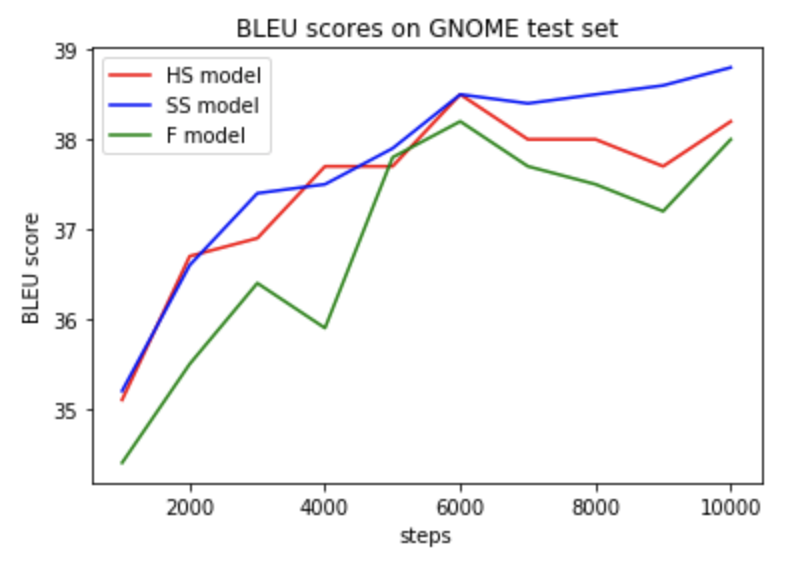}
  \label{fig:sfig3}
\end{subfigure}%
\begin{subfigure}{.25\textwidth}
  \centering
  \includegraphics[width=\linewidth]{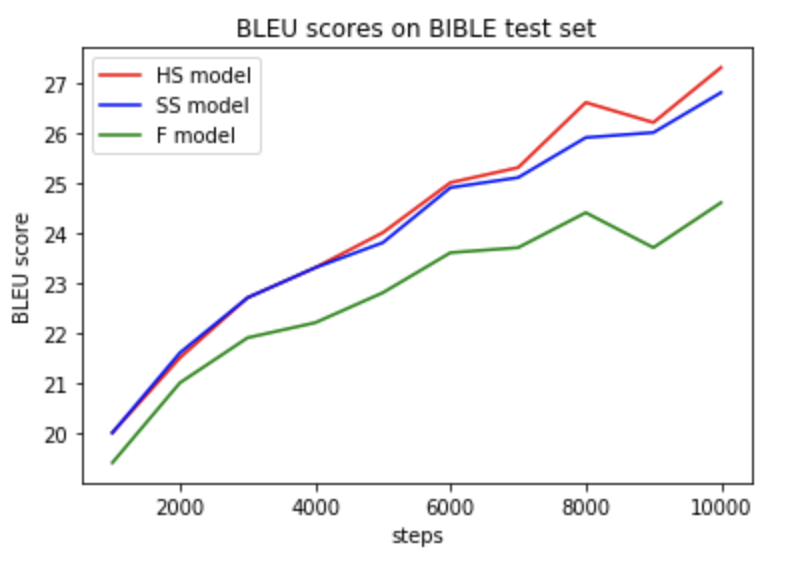}
  \label{fig:sfig4}
\end{subfigure}%
\caption{Evolution of the BLEU score for the configuration: 4 teachers}
\label{fig:fig}
\end{figure}

Moreover, we show in \textbf{Figure \ref{fig:fig}} the  evolution  of  the  BLEU  score  on  the different test sets for the last configuration where 4 teachers are built.

\begin{table}
{\caption{Performance of the teachers, trained with label smoothing, on their corresponding test set.}\label{table3}}
\begin{center}
  \begin{tabular}{| c | c | c |}
\hline
\textbf{Teacher} & \textbf{Domain test set} & \textbf{WMT14 test} \\ \hline 
\hline
Soft teacher Legal & 58.8  & 35.9 \\ \hline
Soft teacher Medical & 53.5 & 35.7 \\ \hline
Soft teacher Software &  39.3 & 29.1 \\ \hline
Soft teacher Religion &  27.4 & 26.8 \\ \hline
  \end{tabular}
\end{center}
\end{table}

If we compare the results shown in Table \ref{table3} with those in Table \ref{table1}, we notice that teachers trained without label smoothing tend to perform better than the ones trained with label smoothing on their specific domain but the opposite is observed on the generic domain. This confirms the assumption that label smoothing results in better generalization.

\begin{table}
{\caption{BLEU scores corresponding to four configurations with 2 to 4 domains using the four different strategies: knowledge distillation with hard teachers, knowledge distillation with soft teachers, mixed finetuning and ensembling.}\label{table4}}
\begin{center}
\begin{adjustbox}{width=\columnwidth,center}
  \begin{tabular}{| c |c |c |c| c |c | c|}
    \hline
    \textbf{Configurations} & \textbf{Model} & \textbf{Legal} & \textbf{Medical} & \textbf{Software} & \textbf{Religion} & \textbf{WMT14}\\ \hline
    Initial Generic Model & - & 51.0 &  33.3& 24.6  & 11.5 & 38 \\ \hline
    \begin{tabular}{@{}c@{}}Legal \\ Medical \end{tabular} & \begin{tabular}{@{}c@{}}HS \\ SS \\ F \\ Ensemble\end{tabular} &\begin{tabular}{@{}c@{}} \textbf{58.4} \\ 58.0 \\  57.4 \\ 57.1 \end{tabular}  & 
      \begin{tabular}{@{}c@{}} \textbf{53.0} \\ 52.0 \\   50.0 \\ 41.1 \end{tabular}& - & -  & \begin{tabular}{@{}c@{}} 35.2 \\ 35.2 \\ 34.4 \\ \textbf{36.9} \end{tabular}\\ \hline

    \begin{tabular}{@{}c@{}}Legal - Medical \\ Software \end{tabular} & \begin{tabular}{@{}c@{}}HS \\ SS \\ F \\ Ensemble \end{tabular} & \begin{tabular}{@{}c@{}}\textbf{57.9} \\ 57.7 \\ 57.0 \\ 54.6 \end{tabular}  & 
     \begin{tabular}{@{}c@{}}\textbf{52.7} \\ 51.4 \\ 50.0 \\ 36.6 \end{tabular}& \begin{tabular}{@{}c@{}}40.2 \\ \textbf{40.6} \\ 40.5\\ 28.6 \end{tabular} & -  & \begin{tabular}{@{}c@{}} 35.4 \\ 35.4 \\ 35.1 \\ \textbf{35.9} \end{tabular}\\ \hline

    \begin{tabular}{@{}c@{}}Legal - Medical \\ Software - Religion \end{tabular} & \begin{tabular}{@{}c@{}}HS \\ SS \\ F \\ Ensemble \end{tabular} & \begin{tabular}{@{}c@{}}\textbf{56.7} \\ 56.5 \\ 56.4 \\ 54.1 \end{tabular} & 
     \begin{tabular}{@{}c@{}} \textbf{51.4} \\ 50.2 \\ 48.8 \\ 34.5 \end{tabular}& \begin{tabular}{@{}c@{}} 37.7 \\ \textbf{38.8} \\ 37.5 \\ 25.4 \end{tabular} & \begin{tabular}{@{}c@{}}\textbf{27.3} \\ 26.8 \\24.4 \\ 15.8 \end{tabular} & \begin{tabular}{@{}c@{}} 34.0 \\ 34.0 \\ 34.9 \\ \textbf{35.8} \end{tabular}\\ \hline
  \end{tabular}
 \end{adjustbox}
\end{center}
 \end{table}
 
Before introducing a metric that will help us evaluate the label smoothing impact, we analyze the results on each dataset.

\begin{itemize}
\item On \textbf{WMT14} all methods result in a decrease of the BLEU score compared to the initial generic model with \textbf{38} BLEU (39.9 with multibleu) as the training goal is to optimize the performance on in-domain test sets.
The decrease amount is approximately similar for all three specialization methods. Ensembling, is generally performing best on this set, but with the a lot more additional complexity and a significant decrease on in-domain data.
\item On \textbf{Legal}, the generic model already provides a quite high  BLEU score : 51. Indeed, the Legal dataset that consists of a collection of legislative documents is quite similar to the training data of the initial generic model.  Even if original BLEU score was high, domain adaptation allows to get even better scores. The HS model outperforms the other models on this domain in all configurations.
\item On \textbf{Medical}, domain adaptation allows to get significant improvement over the initial generic model (up to 20 points in BLEU) as this medical corpus is different from WMT14 data on which the initial generic model was trained. The HS model yields the highest BLEU score in all configurations. 
\item On \textbf{Software}, the improvement is also significant over the initial generic model. The GNOME dataset, added in the configuration with 3 domains, contains short and specific sentences. As it is less specific than Legal and Medical, all three methods perform similarly. 
\item On \textbf{Religion}, the initial generic model results in a low BLEU score. Indeed, the BIBLE dataset contains sentences written in old English/French. On this complex specialization set, the knowledge distillation strategies outperform the mixed finetuning method (gain of 2.65 BLEU in average).   
\end{itemize}

\noindent
We now define the metric $\Delta$ as the average gain over the finetuned model in terms of the BLEU score. 

\noindent
Let $\Delta_{HS}$ and $\Delta_{SS}$ correspond respectively to the $\Delta$ metric computed using the HS model and the SS model. They are defined as following:

\begin{equation}
    \left\{
\begin{array}{l}
  \Delta_{HS} = avg_{datasets}(BLEU_{HS} - BLEU_{F}) \\
  \Delta_{SS} = avg_{datasets}(BLEU_{SS} - BLEU_{F})
\end{array}
\right.
\end{equation}\\

\noindent
We compute both metrics using the results presented in Table \ref{table4} :
\begin{table}
{\caption{Average BLEU gap between knowledge distillation based and finetuning based trained multi-domain models}\label{table5}}
\begin{center}
  \begin{tabular}{| l |c |r | r |}
    \hline
    Configuration & $\Delta_{HS}$ & $\Delta_{SS}$ \\ \hline
    \hline
    2 teachers &  2.0  & 1.3 \\ \hline
    3 teachers &  1.1  & 0.73 \\ \hline
    4 teachers &  1.55 &  1.45 \\ \hline
  \end{tabular}
\end{center}
\end{table}


\noindent
\textbf{Conclusion.}

Our experiments showed that knowledge distillation outperforms finetuning and ensembling in all the tested configurations and confirmed the impact of label smoothing on the distillation process as the student trained using hard teachers performs better than the one that was trained using soft teachers. It also showed that our intuitions about the increasing nature of $\Delta_{HS}$ and $\Delta_{SS}$ aren't true. To our opinion, the distillation strategy is sensitive to the choice of \textbf{$\lambda$}, as it sets up the combination between the negative log-likelihood and the Kullback-Leibler divergence, therefore using the same value of \textbf{$\lambda$} for all the teachers may have limited the potential of this strategy.

\section{Conclusions and Future Directions}

In our work, we showed that knowledge distillation enables to gather the expertise of multiple teachers in one student reducing memory cost and inference time. \\

\noindent
Indeed, we have presented an approach to building a single multi-domain model model incorporating the cognition of multiple single-domain experts with knowledge distillation. We also introduced a state-of-the art method to finetune models on small specialized datasets. Experimental results on English-French translations tasks on Medical, Legal, Software and Religion specialized datasets demonstrate the capability of our approach to outperform finetuning methods while being as scalable and effective as a generic translation model. \\

\noindent
In future work, we plan to explore building a multilingual multi-domain model by distilling multiple unilingual multi-domain models trained using our approach. \\

\noindent
This approach may easily be applied to other domains such as speech-to-text and computer vision as the knowledge transfer is only made through probability distributions.

\section{Acknowledgements}
The authors would also like to thank Mr. \textbf{Pierre Bertrand}, Mr. \textbf{Bruce Delattre}, and Mr. \textbf{Geoffrey Scoutheeten} of BNP Paribas for their valuable comments and suggestions.
The work is supported by the Data \& AI Lab of BNP Paribas.

\bibliography{ecai}
\end{document}